\pdfoutput=1
\documentclass[journal]{IEEEtran}

\usepackage{cite}
\usepackage{amsmath,amssymb,amsfonts}
\usepackage{graphicx}
\usepackage{booktabs}
\usepackage[ruled,linesnumbered]{algorithm2e}
\usepackage{xcolor}
\usepackage[colorlinks=true,allcolors=blue]{hyperref}
\usepackage{url}
\usepackage{tabularx}
\definecolor{ccblue}{RGB}{47,102,168}
\definecolor{ccorange}{RGB}{221,126,55}
\definecolor{ccgreen}{RGB}{60,148,94}
\definecolor{ccpurple}{RGB}{124,86,163}
\definecolor{ccred}{RGB}{198,63,63}
\definecolor{ccslate}{RGB}{78,84,94}

\newcommand{\eg}{e.g.}
\newcommand{\ie}{i.e.}
\newcommand{\etal}{et al.}

\begin{document}

\title{HEALPix-Based Implicit Neural Compression of Omnidirectional Images}

\author{Yulong~Cheng, Youneng~Bao, Junfeng~Zhou, Mu~Li, and~Jie~Wen%
\thanks{Preprint. This work has been submitted to the IEEE for possible
publication. Copyright may be transferred without notice, after which
this version may no longer be accessible.}%
\thanks{\textit{Corresponding author: Mu Li.}}%
\thanks{Y. Cheng, J. Zhou, M. Li, and J. Wen are with the Harbin Institute
of Technology, Shenzhen, China (e-mail: dark.stven@gmail.com;
limu403@163.com). Y. Bao is with Shenzhen University, Shenzhen, China
(e-mail: baoyn@szu.edu.cn).}}

\markboth{Preprint --- submitted to IEEE Transactions on Image Processing}%
{Cheng \MakeLowercase{\textit{et al.}}: HEALPix-Based Implicit Neural Compression of Omnidirectional Images}

\maketitle

% ===========================================================================
\begin{abstract}
Real-time decoding of omnidirectional images is essential for immersive virtual reality (VR) experiences. Compared with conventional transform-based learned image compression (LIC), overfitted codecs offer lower decoding complexity by representing each image with a lightweight implicit neural representation (INR). However, existing approaches typically operate on the equirectangular projection (ERP), which introduces nonuniform spherical sampling and spatial distortion, leading to redundant sampling near the poles and latitude-dependent receptive fields. To address these limitations, we propose a spherical overfitted codec based on a multi-scale Hierarchical Equal Area isoLatitude Pixelization (HEALPix) representation, which enables equal-area sampling and efficient hierarchical modeling of spherical signals. Building on this representation, we develop a Spherical Inter-Scale Conditional entropy model that exploits spatial dependencies and cross-scale correlations among HEALPix latents to improve coding efficiency. We further introduce a Multi-Ring Spherical Convolution that aggregates neighboring pixels over a nearly uniform spherical receptive field across latitudes. Experiments on a public omnidirectional image dataset demonstrate BD-rate reductions of 16.3\%, 26.5\%, and 19.5\% over VVC in terms of V-PSNR, V-MS-SSIM, and V-LPIPS, respectively. Our method also achieves a 13.1\% BD-rate reduction over ERP-based Cool-Chic 5.0 in terms of V-PSNR, while providing $2.6$--$3.6\times$ faster decoding than the evaluated learned autoencoder codecs.
\end{abstract}

\IEEEpeerreviewmaketitle

% ===========================================================================
\section{Introduction}
\label{sec:intro}
\IEEEPARstart{O}{mnidirectional} images, also known as 360\textdegree{} images, capture the complete $360^{\circ}\!\times\!180^{\circ}$ field of view, allowing viewers to explore a scene in any direction. They constitute the primary visual content of virtual reality (VR) and augmented reality (AR) applications, where high spatial resolutions (\eg{}, 8K and beyond) are required to maintain visual quality within the displayed viewport. Such resolutions, however, incur substantial storage and transmission costs. Moreover, decoding takes place on resource-constrained viewing devices, such as head-mounted displays, and must keep pace with viewport rendering to sustain smooth interaction. An omnidirectional image codec must therefore combine high compression efficiency with low decoding complexity.

\begin{figure}[!t]
\centering
\includegraphics[width=\columnwidth]{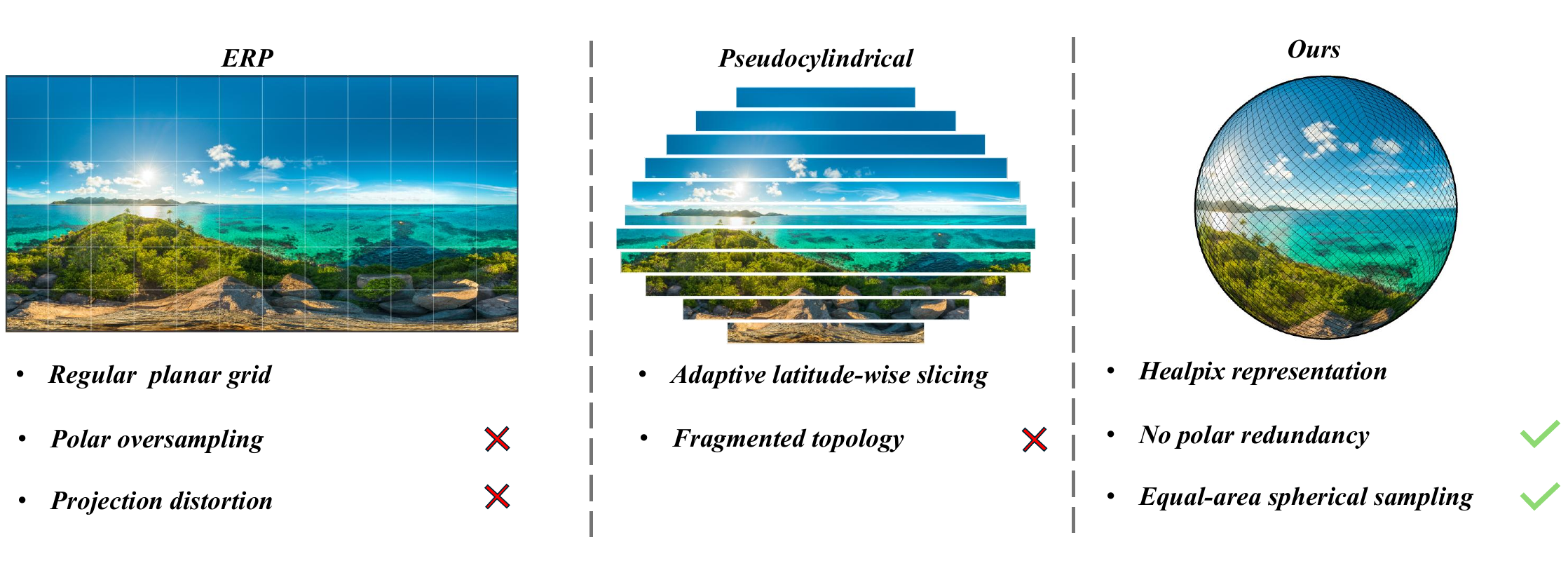}
\caption{Representations of an omnidirectional image, including the ERP, pseudocylindrical representation~\cite{Li2024ContentPseudo} and ours.
}
\label{fig:compare}
\end{figure}

Existing omnidirectional image compression methods predominantly follow a transform-based coding framework. A straightforward approach is to map spherical images onto the equirectangular projection (ERP) and compress them using conventional or learned codecs designed for planar images~\cite{he2018content, xu2020state}. However, ERP introduces nonuniform spherical sampling and projection-induced spatial distortion, resulting in severe oversampling near the poles. To address these issues, existing methods have explored both coding adaptations and alternative spherical representations. For example, LIC360~\cite{Li2022EndToEnd} retains the ERP representation but introduces latitude-adaptive rate allocation to reduce coding redundancy. Alternatively, pseudocylindrical representations~\cite{Li2021Pseudo, Li2024ContentPseudo} adjust the sampling density across latitudes, while HEALPix-based spherical representations~\cite{Wawerek2025OSLOIC, hu2025learned, hu2025feature} enable nearly uniform sampling directly on the sphere. Despite these advances, existing learned omnidirectional image codecs largely rely on transform-based autoencoders trained on large datasets. Their coding performance may degrade under domain shifts~\cite{vanrozendaal2021instance, katakol2021danice, tsubota2023universal}, while their computationally intensive decoders remain a major obstacle to real-time VR applications.

In contrast, overfitted codecs~\cite{COINCW, coin++, HAC, 2023cool, kim2024c3} adopt an instance-specific optimization paradigm. Rather than training a shared model on large datasets, they optimize a compact neural representation for each image and transmit its quantized parameters and, when applicable, latent feature grids in the bitstream. This representation can be viewed as an implicit neural representation (INR)~\cite{SIREN}, which reconstructs image content through a neural function of spatial coordinates. This paradigm offers two key advantages. First, instance-specific optimization allows the coding resources to adapt to individual image content without relying on generalization from a training dataset. Second, lightweight overfitted codecs enable low-complexity decoding, requiring only a few thousand multiplications per pixel~\cite{kim2024c3, ladune2026coolchic5}. These properties make overfitted codecs promising for resource-constrained VR applications.

Despite these advantages, existing overfitted codecs are primarily designed for planar images and typically operate on ERP when applied to omnidirectional content. Their latent grids, interpolation kernels, and entropy context models implicitly assume uniform sampling on a planar domain. We refer to this assumption as the \emph{planar prior}, which is mismatched with the spherical geometry of ERP and leads to two limitations. (i)~\emph{Inefficient bit allocation}: The sampling density of ERP per unit spherical area increases inversely with the cosine of latitude. Consequently, regions beyond $\pm60^{\circ}$ contain one-third of the samples in an ERP-aligned latent grid despite covering only $13.4\%$ of the sphere, potentially wasting coding resources on oversampled polar regions. (ii)~\emph{Latitude-dependent receptive fields}: A fixed-size planar kernel covers progressively smaller spherical areas toward the poles. At a latitude of $88^{\circ}$, its local spherical footprint is approximately $3.5\%$ of that at the equator, leading to inconsistent spatial context across latitudes.

To overcome these limitations, we propose a novel multi-scale spherical INR representation for overfitted omnidirectional image compression, built upon HEALPix grids~\cite{Gorski2005HEALPix}. Fig.~\ref{fig:compare} illustrates our representation in comparison with ERP and the content-weighted pseudocylindrical representation~\cite{Li2021Pseudo}. Our representation offers three key advantages. First, it distributes latent samples uniformly over the sphere, avoiding the polar oversampling inherent in ERP and enabling more balanced allocation of coding resources. Second, it organizes latent grids into a hierarchical structure, allowing image content to be represented and refined across multiple spatial scales. Third, it preserves an iso-latitude ring structure, enabling efficient neighborhood access and structured decoding of spherical latents.
Building on the proposed spherical representation, we develop two modules tailored to its geometry and multi-scale structure. First, the Spherical Inter-Scale Conditional Entropy Model exploits both spatial and cross-scale dependencies to improve coding efficiency. Specifically, it selects previously decoded neighbors at the same scale based on spherical distance rather than a fixed planar context mask, and further conditions each latent on its ancestors at coarser scales. This design enables spherical geometry-aware context modeling without introducing latitude-dependent spatial neighborhoods. Second, the Multi-Ring Spherical Convolution aggregates features from all pixels within a specified number of hops on the spherical grid, maintaining a nearly uniform spherical receptive field across latitudes. The resulting codec is optimized separately for each image, with distortion measured by the mean squared reconstruction error over a dense set of approximately uniformly distributed points on the sphere. In experiments, since VR users observe rectilinear viewports rather than ERP images, we evaluate the reconstruction quality of all codecs on the displayed viewports.
Compared with H.266/VVC~\cite{VVC}, our codec reduces the bit rate by $16.3\%$, $26.5\%$, and $19.5\%$ under V-PSNR, V-MS-SSIM, and V-LPIPS, respectively, outperforms all compared codecs, including MLIC++~\cite{jiang2023mlicpp} and OSLO-IC, and decodes $2.6$--$3.6\times$ faster than the compared learned autoencoders on a single CPU core.

The main contributions of this paper are summarized as follows:
\begin{itemize}
\item \textbf{Multi-scale spherical INR representation.} We propose a novel multi-scale spherical INR representation for overfitted omnidirectional image compression. Built upon equal-area HEALPix grids, our representation avoids the polar oversampling of ERP and organizes spherical latents hierarchically, enabling balanced spatial sampling and efficient multi-scale modeling.

\item \textbf{Spherical geometry-aware coding modules.} We develop a Spherical Inter-Scale Conditional Entropy Model that exploits spatial dependencies through spherical distance-based context selection and cross-scale correlations through hierarchical conditioning. We further introduce a Multi-Ring Spherical Convolution that maintains a nearly uniform spherical receptive field across latitudes, enabling geometry-consistent feature extraction.

\item \textbf{Viewport-based evaluation and improved coding efficiency.} We establish a viewport-based evaluation protocol that measures reconstruction quality along identical viewing rays, using actual bitstream rates and evaluation samples distinct from those used for fitting. Experiments demonstrate a 13.1\% BD-rate reduction over Cool-Chic 5.0 in terms of V-PSNR, with $2.6$--$3.6\times$ faster decoding than the evaluated learned autoencoders.
\end{itemize}

The remainder of this paper is organized as follows.
Section~\ref{sec:related} reviews related work,
Section~\ref{sec:method} presents the proposed codec,
Section~\ref{sec:experiments} reports the experimental results, and
Section~\ref{sec:conclusion} concludes the paper.
% ===========================================================================
\section{Related Work}
\label{sec:related}

In this section, we give a brief overview of learned image compression
and of INR-based codecs, and then review compression methods for
omnidirectional images.

\subsection{Learned Image Compression}

Learned image codecs follow the nonlinear transform coding
paradigm~\cite{balle2017endtoend}: an image is mapped by learned
analysis and synthesis transforms, and the resulting latents are
quantized and compressed by entropy coding. 
Since the hyperprior model~\cite{balle2018variational}, subsequent work has mainly
improved the transforms and the entropy model. For the entropy model,
autoregressive context models~\cite{minnen2018joint, Cheng2020} improve
compression efficiency but decode serially, and ELIC~\cite{elic} groups
the latents along space and channel to reduce this cost.
MLIC++~\cite{jiang2023mlicpp} combines channel-wise, local, and global
spatial contexts at linear complexity. For the transforms, TCM~\cite{liu2023tcm}
combines Transformer and CNN branches, FTIC~\cite{li2024ftic} uses
frequency-decomposition window attention, and
LALIC~\cite{feng2025lalic}, MambaIC~\cite{zeng2025mambaic}, and
CMIC~\cite{chen2026cmic} replace quadratic attention with linear
attention or content-aware state space models. Other work improves
training, for example by moving informative content into earlier
decoded groups~\cite{han2024cca} or by assigning different frequency
bands to different scales~\cite{cong2026taming}.

All of these codecs are trained on large datasets, so their decoders
must generalize across images. This has two consequences. First, the decoders are computationally demanding: modern autoencoders such as
MLIC++ require on the order of one million multiplications to decode
each pixel~\cite{ladune2026coolchic5}. Second, a decoder trained to
perform well on average over a dataset is not optimal for any
particular image, and adapting the codec to the content of each image
improves its rate--distortion
performance~\cite{vanrozendaal2021instance, katakol2021danice,
tsubota2023universal}.

\subsection{INR-Based Image Compression}
Early overfitted codecs emerged from INRs~\cite{SIREN}, which represent
a signal as a function modeled by a neural network. COIN~\cite{COINCW}
applies this idea to image compression: it overfits a sinusoidal MLP
mapping pixel coordinates to RGB values, and transmits the network
itself as the compressed representation. Subsequent works speed up the
fitting with Fourier features~\cite{Fourier} and meta-learned
initializations~\cite{HAC}. However, a single coordinate MLP is biased towards low-frequency
information~\cite{rahaman2019spectral}, which makes it difficult to represent
high-frequency details, and its few parameters must describe the entire
image. Other works refine specific components of
these codecs. COMBINER~\cite{guo2023combiner} compresses the network
weights with relative entropy coding, RQAT-INR~\cite{RQAT-INR} improves
quantization-aware training, TINC~\cite{tinc} partitions the signal
with a tree structure, and perceptual quality is optimized with a
Wasserstein distortion~\cite{Wasserstein-INR}.

Cool-Chic~\cite{2023cool} overcomes the limited capacity of a single
MLP by introducing hierarchical latent grids, which build on
multi-resolution feature grids~\cite{muller2022instantngp}. The latent
grids are decoded by a lightweight network and entropy coded with an
autoregressive probability model. Its follow-up
works~\cite{leguay2023lowcomplexity, kim2024c3, ladune2026coolchic5}
require only a few hundred to a few thousand multiplications per
decoded pixel, while offering rate--distortion performance comparable
to VVC~\cite{VVC}. Our work extends these hierarchical latent grids to
the sphere by defining them on HEALPix.

INRs are also applied to signals beyond planar images. Generalized
INRs~\cite{grattarola2022giner} extend them to signals on graphs and
manifolds. HNeR-S~\cite{kim2024hners} fits HEALPix feature grids to
weather and cosmic microwave background maps, and measures compression
through the parameter count. For 360\textdegree{} video,
NeRV360~\cite{nerv360} fits a network to ERP frames and decodes the
viewports directly, without a rate--distortion pipeline.
MV-HiNeRV~\cite{kwan2024mvhinerv} compresses multi-view video, which is
not omnidirectional. Yet, none of these methods is a codec on the sphere that combines
entropy coding with rate--distortion optimization.

\subsection{Omnidirectional Image Compression}
\label{sec:related_omni}
Omnidirectional images are usually projected onto a plane, most often
with ERP, and then compressed with a standard
codec~\cite{he2018content, xu2020state}. In streaming, viewport-adaptive
schemes further adjust the rate allocation~\cite{yaqoob2020survey}.
The resulting mismatch between the plane and the sphere has been
addressed in three ways.

The first approach changes the projection. Cubemap projection (CMP)
and its variants~\cite{he2018content} reduce the average distortion of
ERP, and pseudocylindrical projections~\cite{Youvalari2016Pseudo}
reduce the stretching near the poles. Encoder optimization, rate
control, and rate--distortion optimization have also been adapted to
spherical geometry~\cite{Li2017SphericalRDO}. However, CMP introduces discontinuities at face boundaries, and the
grids of these and similar projections still do not sample the sphere
uniformly.

The second approach keeps ERP or CMP, but makes the network or the loss
aware of the sphere. This is done through latitude-dependent
convolutions~\cite{su2017learning}, adjusted kernel sampling
locations~\cite{Coors2018SphereNet}, or tangent-image
decomposition~\cite{eder2020tangent}. For learned compression, an
end-to-end optimized 360\textdegree{} codec~\cite{Li2022EndToEnd}
combines latitude-adaptive bit allocation with a viewport-based
distortion. Pseudocylindrical representations~\cite{Li2021Pseudo}
sample the sphere nearly uniformly and are processed by standard
convolutions with pseudocylindrical padding, and their content-weighted
version~\cite{Li2024ContentPseudo} adapts the strip layout to each
image. Yet, these methods still operate on 2D grids, so the mismatch
between the plane and the sphere is reduced but not removed.

The third approach avoids projection altogether. Spherical
CNNs~\cite{Cohen2018SphericalCNN, esteves2018so3} are
rotation-equivariant by design, but require repeated spherical harmonic
transforms. Icosahedral and mesh-based
methods~\cite{cohen2019gauge, lee2019spherephd} are more efficient, but
give up equivariance, and their cells have only approximately equal
area. HEALPix~\cite{Gorski2005HEALPix} instead divides the sphere into
pixels of exactly equal area at every resolution, so that the same bits
per pixel give the same bits per unit area. It has been adopted for
CNNs on the sphere~\cite{Perraudin2019DeepSphere} and for learned
compression. OSLO~\cite{Bidgoli2022OSLO} redefines the convolution,
downsampling, and upsampling of an autoencoder on HEALPix, and
OSLO-IC~\cite{Wawerek2025OSLOIC} adds attention modules and a spatial
context model. The transformer-based HEALPix
codecs~\cite{hu2025learned, hu2025feature} rely on windowed spherical
self-attention, cross-window spherical convolution, and a transformer
context model. They start from the same observation as our multi-ring
convolution, \ie{}, a fixed nine-pixel HEALPix kernel has a small
receptive field. Although these codecs remove the projection, they
remain autoencoders trained on large datasets, with large decoders.

Our codec is an overfitted codec on the
HEALPix sphere, in which the latent grids, the entropy model, and the
synthesis network are all defined on equal-area pixels, and its decoder
has only a few thousand weights. Since this decoder is rebuilt from the
bitstream alone and every weight costs bits, our modules are designed
to be geometrically exact at almost no cost: the multi-ring convolution
grows its receptive field ring by ring, and the inter-scale entropy
model takes its context from the HEALPix quad-tree itself.

% ===========================================================================

\section{Proposed Method}
\label{sec:method}

\begin{figure*}[!t]
    \centering
    \includegraphics[width=\textwidth]{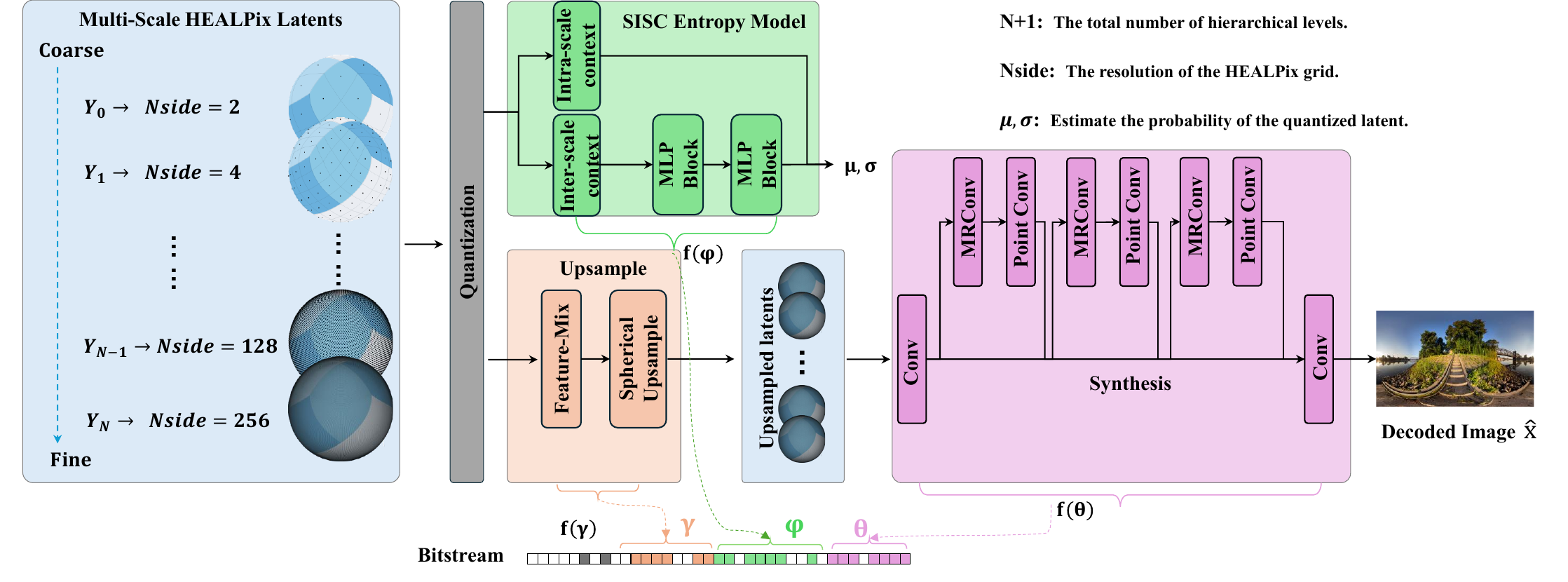}
    \caption{\textbf{Overall architecture.} Each panorama is represented
by hierarchical latent grids $\{Y_i\}_{i=0}^{N}$ on the HEALPix sphere
($N_{\mathrm{side}}\in\{2,4,\ldots,256\}$), which are quantized and
fitted to the image. Top: the SISC entropy model $f_\varphi$ predicts the Laplace parameters of each latent from same-level and ancestor contexts (Fig.~\ref{fig:entropy}) for the arithmetic coder. Bottom: the
spherical upsampler $f_\gamma$ merges the levels from coarse to fine, and
the multi-ring synthesis network $f_\theta$ (Fig.~\ref{fig:ring})
reconstructs the sphere. The latents and the decoder weights are trained with SOAP on
$D+\lambda\,\mathcal{R}_{\mathrm{latent}}/(HW)$; the weight rate
$\mathcal{R}(\varphi,\gamma,\theta)$ of Eq.~\eqref{eq:rd_loss} is not
differentiated and is handled by a rate--distortion search over the
weight quantization at the end of fitting
(Sec.~\ref{sec:objective}).}
    \label{fig:pipeline}
\end{figure*}

In this section, we first present the spherical latent representation of
our codec (Sec.~\ref{sec:representation}) and its quantization
(Sec.~\ref{sec:quant}). We then describe the Spherical Inter-Scale Conditional (SISC) entropy model that estimates the rate of the latents (Sec.~\ref{sec:entropy}),
the decoder that reconstructs the image from the quantized latents
(Sec.~\ref{sec:decoder}), and finally the rate--distortion optimization
that fits all parameters to each image (Sec.~\ref{sec:objective}). The overall architecture is shown in
Fig.~\ref{fig:pipeline}.

% ---------------------------------------------------------------------------
\subsection{Spherical Latent Representation}
\label{sec:representation}

Given an omnidirectional image $\mathbf{x}$ stored in ERP with height $H$
and width $W$, we map it onto the unit sphere, and denote by
$\mathbf{x}(\mathbf{s})$ its color at direction $\mathbf{s}$.

Each panorama is represented by $N{+}1$ hierarchical latent grids
$\{Y_i\}_{i=0}^{N}$ defined on HEALPix grids. The $i$-th grid has
resolution $N_{\mathrm{side},i}=N_{\mathrm{side},0}\,2^{i}$ and divides
the sphere into $N_{\mathrm{pix},i}=12N_{\mathrm{side},i}^{2}$ sampling
regions, each represented by the sampling point at its center.
$Y_{i,j}$ denotes the single-channel latent at sampling point $j$ of
level $i$, with $j=0,\dots,N_{\mathrm{pix},i}-1$ in RING order, \ie{},
along circles of constant latitude. We use eight levels ($N=7$), with
$N_{\mathrm{side},0}=2$ in the main experiments and
$N_{\mathrm{side},0}=4$ in the high-resolution experiment. The coarse
levels capture the global, low-frequency structure of the panorama,
while the finer levels progressively add local texture. The latent
grids are trainable parameters that represent the image.

Two geometric properties of HEALPix determine the spatial distribution
and the hierarchy of the latent representation.

\begin{table}[!tb]
\centering
\caption{Relative standard deviation (\%) of neighbor distances in each
direction over the sphere, for ERP at $768\times1536$ and HEALPix at
$N_{\mathrm{side}}=256$. Lower is more uniform; the best mean is in
bold.}
\label{tab:sampling}
\footnotesize
\setlength{\tabcolsep}{2pt}
\resizebox{\columnwidth}{!}{%
\begin{tabular}{lccccccccc}
\toprule
Projection & SW & W & NW & N & NE & E & SE & S & \textbf{Mean} \\
\midrule
ERP     & 11.99 & 48.26 & 11.99 & 0.00 & 11.99 & 48.26 & 11.99 & 0.00 & 18.06 \\
HEALPix & \textbf{11.15} & \textbf{10.28} & \textbf{11.15} & 13.83 & \textbf{11.15} & \textbf{10.28} & \textbf{11.15} & 13.83 & \textbf{11.60} \\
\bottomrule
\end{tabular}}
\end{table}

Property 1 (equal area): Ideally, the samples of an omnidirectional
image should be distributed uniformly over the sphere, and the number of
bits spent in each region varies with the image content rather than with
the latitude. On HEALPix, all sampling regions of a level cover the same
area on the sphere and the grid itself causes no redundant rate near the
poles. HEALPix also samples the sphere more uniformly than ERP: the
distances between neighboring sampling points have a relative standard
deviation of $11.6\%$, against $18.1\%$ for ERP, as shown in
Table~\ref{tab:sampling}. HEALPix thus naturally satisfies this ideal.
In contrast, an ERP grid samples the sphere much more densely near the
poles than near the equator. At latitude $\delta$\footnote{Throughout
the paper, $\delta$ denotes latitude and $\varphi$ denotes the
entropy-model parameters.}, its areal sampling density,
defined as the number of samples per unit area on the sphere, is
\begin{equation}
    \eta_{\mathrm{ERP}}(\delta) = \frac{HW}{2\pi^{2}\cos\delta}.
    \label{eq:erp_density}
\end{equation}
This density grows without bound toward the poles, whereas it is
constant for HEALPix. A planar-grid codec thus places its latents with a
density proportional to $\sec\delta$, \eg{}, the rows with
$|\delta|>60^{\circ}$ hold one third of all ERP latents but cover only
$13.4\%$ of the sphere.

Property 2 (exact quad-tree): Each refinement splits a sampling region
into four equal-area regions that exactly cover it. The levels thus form
a quad-tree, in which every sampling point has a unique parent at the
next coarser level, namely the point whose region contains it. We write
$\mathrm{par}(j)$ for the parent of sampling point $j$ and
$\mathrm{par}^{k}(j)$ for its ancestor $k$ levels up. This parent--child
relation gives an exact correspondence between levels.

% ---------------------------------------------------------------------------

\begin{figure*}[!t]
    \centering
    \includegraphics[width=\textwidth]{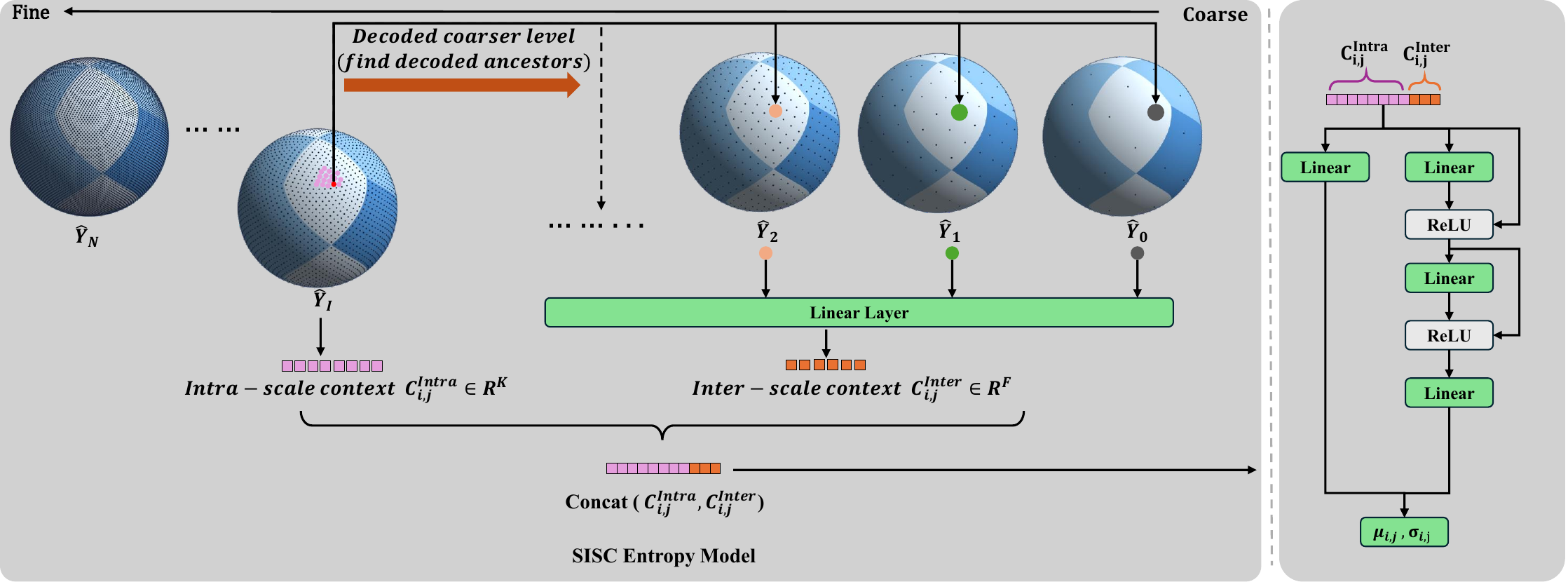}
    \caption{\textbf{Spherical Inter-Scale Conditional (SISC) Entropy Model.}
    Left: for a latent $\hat{Y}_{i,j}$ (red dot), the intra-scale context
    collects its $K$ nearest decoded neighbors (pink), and the inter-scale
    context is obtained by $\mathrm{Linear}_i$ from its decoded ancestors on
    the coarser levels (colored dots). Right: the shared predictor
    $f_\varphi$ maps the concatenated contexts to the Laplace parameters
    (Eq.~\eqref{eq:predictor}).}
    \label{fig:entropy}
\end{figure*}
\subsection{Quantization}
\label{sec:quant}
All transmitted parameters must be quantized, and we quantize the
latents and the decoder weights differently.

\textbf{Latents.} At the end of the fitting, the latents are quantized
by rounding to the nearest integer,
\begin{equation}
  \hat{Y}_{i,j}=\lfloor Y_{i,j}\rceil .
  \label{eq:latent_quant}
\end{equation}
Since rounding has zero gradient almost everywhere, it is replaced
during fitting by a differentiable approximation. Early in the fitting,
this approximation is soft rounding~\cite{agustsson2020uq, kim2024c3}
with additive noise, and both are gradually reduced so that the latents
approach integers. Later, the latents are rounded by
Eq.~\eqref{eq:latent_quant} with a straight-through gradient.

\textbf{Decoder weights.} Each weight $\omega$ of module
$\alpha\in\{\varphi,\gamma,\theta\}$ is quantized with a module-wise step
$\Delta_\alpha$,
\begin{equation}
  \hat{\omega}=\Delta_\alpha\lfloor \omega/\Delta_\alpha\rceil ,
  \label{eq:weight_quant}
\end{equation}
and the integer $\lfloor \omega/\Delta_\alpha\rceil$ is written with an
Exp-Golomb code. The weights are kept in floating point during training.
After training, the step $\Delta_\alpha$ and the order of the
Exp-Golomb code of each module are selected by a rate--distortion search
(Sec.~\ref{sec:objective}), and the weights are quantized accordingly.

% ---------------------------------------------------------------------------

\subsection{Spherical Inter-Scale Conditional Entropy Model}
\label{sec:entropy}
To better exploit the previously decoded latents, the SISC entropy model combines two causal contexts, an intra-scale context formed by the decoded neighbors on the same grid and an inter-scale context formed by the decoded ancestors on the coarser grids. At the same time, we make two geometric adaptations to compute these contexts on the sphere: the neighbors are selected by their distance on the sphere rather than by a fixed context mask on the planar grid, and the ancestors are found through the HEALPix quad-tree. From these contexts, a shared predictor estimates a single Laplace distribution for each quantized latent. Fig.~\ref{fig:entropy} illustrates the model, and we describe the two contexts, the shared predictor, and the probability model in turn.

\subsubsection{Intra-Scale Context}
\label{sec:intra_ctx}
Neighboring latents on the same grid are strongly correlated, and we let
the intra-scale context exploit this local redundancy through the
neighbors that precede the target in the decoding order. On ERP, such
neighbors are usually taken from a fixed context mask on the planar
grid, whose region on the sphere shrinks toward the poles. We instead
select the neighbors by their distance on the sphere, so that the
context covers a region of similar size at every latitude. For a target
latent, the $K$ already decoded sampling points on the same grid that
are nearest to it on the sphere are selected, and their values form the
intra-scale context $\mathbf{C}^{\mathrm{Intra}}_{i,j}$. The first few
sampling points of each level have fewer than $K$ decoded neighbors, and the missing entries are set to zero.

\subsubsection{Inter-Scale Context}
\label{sec:inter_ctx}
Since the latent grids are decoded from the coarsest level $\hat Y_0$ to the
finest level $\hat Y_N$, all coarser levels are already known when a level is
decoded. The inter-scale context therefore consists of the same region
of the sphere at these lower resolutions, which serves as a prior for
that region at the current resolution.

On HEALPix, this cross-scale correspondence can be expressed exactly
through the geometric nesting of the grids. Through the
quad-tree of Property~2, every fine latent lies inside a unique chain of ancestors, one at each coarser level, and these ancestors are exactly the coarse latents that cover its region. As shown in the left part of Fig.~\ref{fig:entropy}, for the latent at position $j$ of level $i$, we find its decoded ancestors on $\hat Y_{i-1},\dots,\hat Y_0$, gather them from coarse to fine, and map them to the inter-scale context with a linear layer:
\begin{equation}
\begin{aligned}
  \mathbf{r}_{i,j} &= \bigl[\hat Y_{0,\mathrm{par}^{i}(j)},\,\dots,\,
  \hat Y_{i-1,\mathrm{par}(j)}\bigr],\\
  \mathbf{C}^{\mathrm{Inter}}_{i,j} &= \mathrm{Linear}_i(\mathbf{r}_{i,j}),
\end{aligned}
\label{eq:ancestors}
\end{equation}
where $\mathrm{Linear}_i$ is a linear layer without activation specific
to level $i$, which maps $\mathbf{r}_{i,j}$ to an $F$-dimensional
context, so that each level can weight its ancestors differently.
The coarsest level has no ancestors, and its inter-scale context is set
to zero. We analyze the effect of this context on the rate of each level in Sec.~\ref{sec:ablation} (Fig.~\ref{fig:rate_maps}).

\subsubsection{Shared Predictor}
\label{sec:predictor}
The two contexts describe a latent from two different views: the
intra-scale context gives its local neighborhood on the same grid, and
the inter-scale context gives the coarser structure of the same region.
A shared predictor combines them and predicts the parameters of a
Laplace distribution:
\begin{equation}
    (\mu_{i,j},\rho_{i,j})
    = f_\varphi\bigl(\mathrm{Concat}(\mathbf{C}^{\mathrm{Intra}}_{i,j},
      \,\mathbf{C}^{\mathrm{Inter}}_{i,j})\bigr),
    \label{eq:predictor}
\end{equation}
where $\mathrm{Concat}$ denotes channel-wise concatenation, $\mu_{i,j}$ is the location, and $\rho_{i,j}$ a raw log-scale, from which the scale $\sigma_{i,j}$ is obtained by a shifted and clipped exponential. As shown in the right part of Fig.~\ref{fig:entropy}, the predictor is composed of two parallel branches: a linear layer and
an MLP with ReLU activations and residual connections. The parameters of $f_\varphi$ and $\mathrm{Linear}_i$ form $\varphi$.

\subsubsection{Single-Laplace Model}
\label{sec:laplace}
Each quantized latent is modeled by a single Laplace distribution with
the predicted location and scale. Since the latents are integers, its
probability is the mass of this distribution over the quantization bin
of the latent:
\begin{equation}
    P\bigl(\hat{Y}_{i,j}\,;\,\mu_{i,j},\sigma_{i,j}\bigr)
    = \int_{\hat{Y}_{i,j}-\frac12}^{\hat{Y}_{i,j}+\frac12}
      \frac{1}{2\sigma_{i,j}}
      \exp\!\Bigl(-\frac{|\xi-\mu_{i,j}|}{\sigma_{i,j}}\Bigr)\,d\xi .
    \label{eq:laplace}
\end{equation}
The latent rate is the total code length of the quantized latents:
\begin{equation}
\mathcal{R}_{\mathrm{latent}}
= \sum_{i=0}^{N}\sum_{j=0}^{N_{\mathrm{pix},i}-1}
  -\log_2 P\bigl(\hat{Y}_{i,j};\mu_{i,j},\sigma_{i,j}\bigr).
\label{eq:latent_rate}
\end{equation}

% ---------------------------------------------------------------------------
\subsection{Decoder: From Latents to the Image}
\label{sec:decoder}
The decoder first decodes the quantized latents
$\{\hat Y_i\}_{i=0}^{N}$ from the bitstream with the SISC entropy model
$f_\varphi$, and then reconstructs the image as RGB values on the finest HEALPix grid:
\begin{equation}
  \hat{\mathbf{x}} = f_\theta\bigl(f_\gamma(\{\hat Y_i\}_{i=0}^{N})\bigr).
  \label{eq:codec}
\end{equation}
We describe the decoder in three parts. We first define the Multi-Ring
Spherical Convolution, the spatial operator used in both networks. We then present the spherical upsampler $f_\gamma$, which merges all levels into a dense representation on the finest grid, and finally the synthesis network $f_\theta$, which maps this representation to RGB values.

\subsubsection{Multi-Ring Spherical Convolution}
\label{sec:multiring}
\begin{figure}[!t]
\centering
\includegraphics[width=\columnwidth]{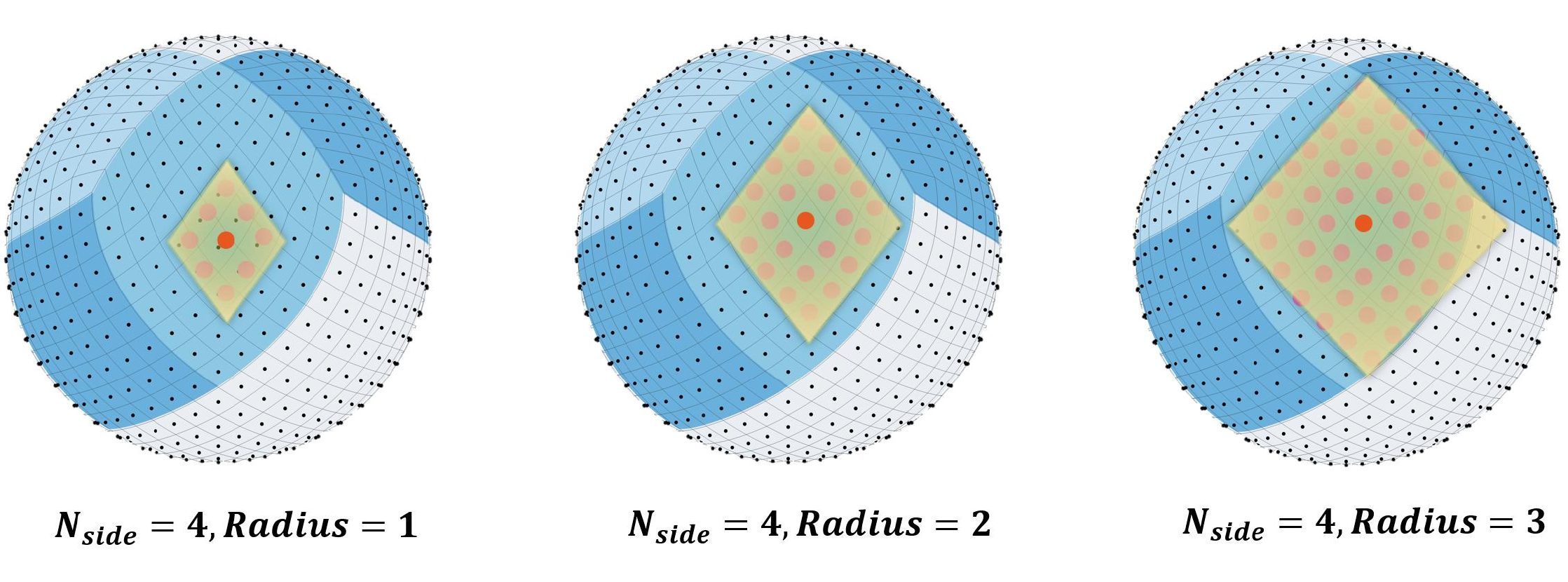}
\caption{\textbf{Multi-Ring Spherical Convolution.} Receptive fields of
radius $R\in\{1,2,3\}$ on the HEALPix $8$-neighbor graph
($N_{\mathrm{side}}{=}4$). Each neighborhood $B_R(v)$ contains every
sampling point within $R$ hops of the center (highlighted).}
\label{fig:ring}
\end{figure}

The fixed nine-point HEALPix neighborhood provides only a small
receptive field~\cite{hu2025learned, hu2025feature}. Windowed spherical
attention enlarges it at a high cost in computation and storage, and
dilated convolution~\cite{yu2017dilated} samples points that move away
from the target and become less correlated with it as the dilation rate
grows. We instead propose a Multi-Ring Spherical Convolution, whose
receptive field covers every sampling point within $R$ hops of the
center, so that the number of kernel weights grows with the covered
area (Fig.~\ref{fig:ring}).

With $\mathrm{hop}(v,q)$ the shortest-path length between sampling
points $v$ and $q$ on the 8-neighbor graph, the $r$-th ring and the
neighborhood of radius $R$ are
\begin{equation}
\begin{gathered}
  S_r(v) = \bigl\{\, q \;\big|\; \mathrm{hop}(v,q)=r \,\bigr\},\\
  B_R(v) = \bigcup_{r=0}^{R} S_r(v).
\end{gathered}
\label{eq:ring_shell}
\end{equation}
This neighborhood contains $(2R{+}1)^2$ sampling
points\footnote{Except near the adjacency defects, \ie{}, the $24$
sampling points with only seven neighbors.}, \eg{}, $49$ for $R=3$, whereas a dilated kernel always
samples $9$. The convolution is depthwise: its response at $v$ is the
weighted sum of the input features over $B_R(v)$, with a separate weight
for each sampling point given by its position in the kernel.

\subsubsection{Spherical Upsampling}
\label{sec:upsampling}
After entropy decoding, the spherical upsampler $f_\gamma$ maps the
hierarchical latent grids to a dense representation on the finest
HEALPix grid, which is then fed to the synthesis network. The grids are
processed from the coarsest to the finest level, and each step consists
of a feature mixing operation and an upsampling operation.

\textbf{Lite feature mixing.} Before upsampling, the quantized grid
$\hat Y_i$ of each level is refined by a small residual block. It
consists of a depthwise multi-ring convolution of radius one, followed by a GELU activation and a pointwise convolution. The output of the block is scaled by a learnable scalar and added back to its input, which gives the mixed features $\bar Y_i$ of level $i$.

\begin{figure}[!t]
\centering
\includegraphics[width=\columnwidth]{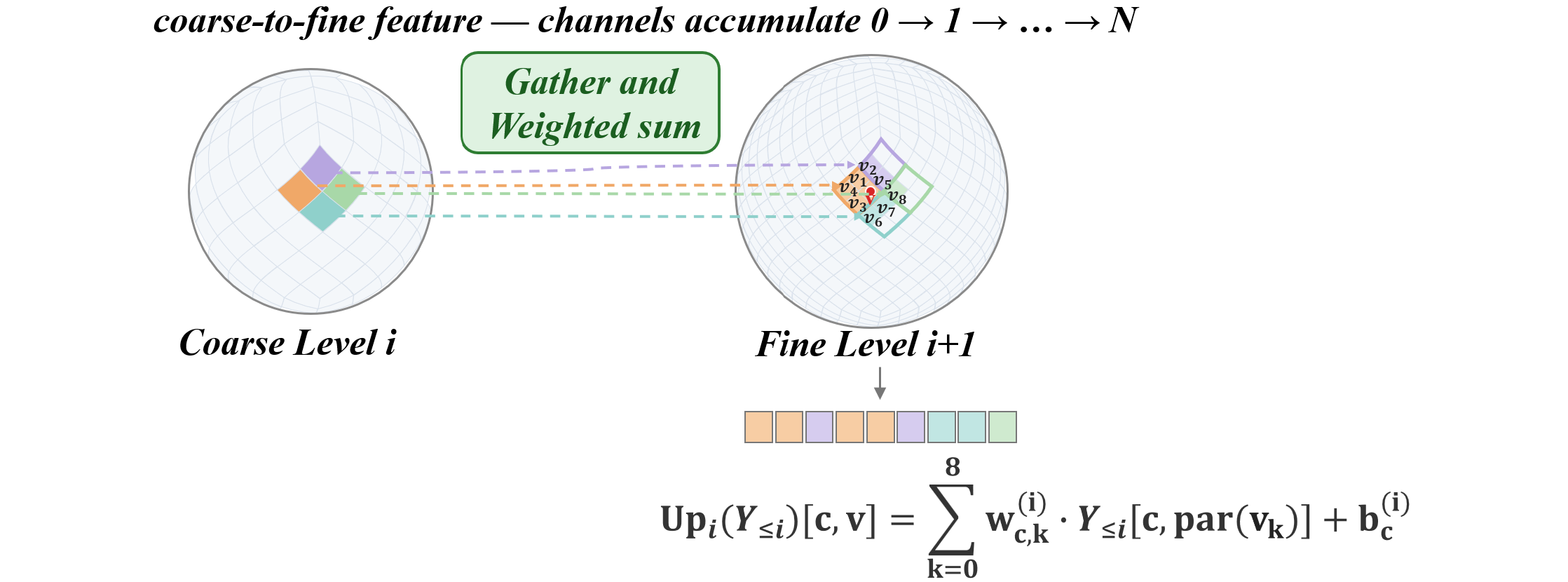}
\caption{Spherical nearest-parent upsampling from level $i$ to level
$i{+}1$. The fine sampling point $v=v_0$ and its eight neighbors
$v_1,\dots,v_8$ are mapped to their parents at level $i$, and the region
of each fine sampling point is drawn in the color of its parent.}
\label{fig:upsample}
\end{figure}

\textbf{Spherical nearest-parent upsampling.} We denote by $Y_{\le i}$
the feature stream that merges levels $0,\dots,i$ on the grid of level
$i$, starting from $Y_{\le 0}=\bar Y_0$. For $i=0,\dots,N-1$, the stream
$Y_{\le i}$, which has $i{+}1$ channels, is upsampled to level $i{+}1$
by the upsampling module $\mathrm{Up}_i$ and concatenated with the mixed
features of that level. As illustrated in Fig.~\ref{fig:upsample}, each
fine sampling point $v=v_0$ and its eight neighbors $v_1,\dots,v_8$ are
sent to their parents at level $i$, and the output of $\mathrm{Up}_i$
at $v$ is a depthwise weighted sum of these parent features:
\begin{equation}
\begin{aligned}
    \mathrm{Up}_i(Y_{\le i})[c,v] &= \sum_{k=0}^{8} w^{(i)}_{c,k}\cdot
    Y_{\le i}[c,\,\mathrm{par}(v_k)] + b^{(i)}_c,\\
    Y_{\le i+1} &= \mathrm{Concat}\bigl(\mathrm{Up}_i(Y_{\le i}),\,\bar Y_{i+1}\bigr),
\end{aligned}
\label{eq:fused_upsample}
\end{equation}
where $c=0,\dots,i$ indexes the channels, $\mathrm{Concat}$ denotes channel-wise
concatenation, and a missing neighbor at an adjacency defect is replaced by $v$. The upsampling filters $w^{(i)}_{c,k}$ and biases $b^{(i)}_c$ are not shared across levels. After the last step, the dense representation $Y_{\le N}$ has
$N{+}1$ channels, where channel $c$ carries the mixed features of level
$c$ upsampled to the finest grid. The upsampling filters and the per-level mixers form $\gamma$.

\subsubsection{Synthesis Network}
\label{sec:synthesis}
The synthesis network $f_\theta$ turns the dense representation
$Y_{\le N}$ into RGB colors by aggregating spatial context on the
sphere. A pointwise convolution $\mathrm{PW}_{\mathrm{in}}$ first
projects $Y_{\le N}$ to the synthesis features $\mathbf{a}^{(0)}$. For $t=1,2,3$, a residual block built on the multi-ring convolution with
radius $R_t=t$ then refines these features, and a final pointwise
convolution $\mathrm{PW}_{\mathrm{out}}$ maps them to the RGB values
$\hat{\mathbf{x}}$:
\begin{equation}
\begin{aligned}
    \mathbf{a}^{(0)} &= \mathrm{PW}_{\mathrm{in}}(Y_{\le N}),\\
    \mathbf{a}^{(t)} &= \mathbf{a}^{(t-1)} + \mathrm{GELU}\bigl(\mathrm{PW}_t(\mathrm{MRConv}_{R_t}(\mathbf{a}^{(t-1)}))\bigr),\\
    \hat{\mathbf{x}} &= \mathrm{PW}_{\mathrm{out}}(\mathbf{a}^{(3)}),
\end{aligned}
\label{eq:synth_block}
\end{equation}
where $\mathbf{a}^{(t)}$ denotes the output of the $t$-th block,
$\mathrm{MRConv}_{R_t}$ the depthwise multi-ring convolution of radius
$R_t$, and $\mathrm{PW}_t$ a pointwise convolution. In each block, the
multi-ring convolution gathers spatial context on the sphere, and the
pointwise convolution mixes the channels. As the radius grows from
block to block, the receptive field widens from the nearest neighbors to
a larger region of the sphere, while the residual connections keep the
local details. The parameters of these layers form $\theta$.
% ---------------------------------------------------------------------------
\begin{algorithm}[!t]
\caption{Per-image fitting}
\label{alg:training}
\KwIn{ERP image $\mathbf{x}$; points $\{\mathbf{s}_m\}$ with colors
$\mathbf{x}(\mathbf{s}_m)$ interpolated once from $\mathbf{x}$}
\KwOut{Bitstream containing the quantized decoder weights
$\varphi,\gamma,\theta$ and the latents $\{\hat{Y}_i\}_{i=0}^{N}$ coded
with the entropy model}
Load the lookup tables of the entropy model, the spherical upsampler,
the multi-ring convolution, and the readout\;
\tcp{Stage 1: soft rounding with decreasing noise}
\For{\textnormal{each iteration}}{
  Soft-round the latents with additive noise\;
  Estimate $\mathcal{R}_{\mathrm{latent}}$ by Eq.~\eqref{eq:latent_rate}\;
  Decode the sphere with $f_\gamma$ and $f_\theta$, and compute $D$ by
  Eq.~\eqref{eq:distortion}\;
  Update all parameters with SOAP on
  $D+\lambda\,\mathcal{R}_{\mathrm{latent}}/(HW)$, and reduce the
  soft rounding and the noise\;
}
\tcp{Stage 2: straight-through rounding}
\For{\textnormal{each iteration}}{
  Round the latents with a straight-through gradient, and perform the
  same update as in stage~1\;
}
\tcp{Weight quantization}
Select the step $\Delta_\alpha$ and the Exp-Golomb order of each module
$\alpha\in\{\varphi,\gamma,\theta\}$ by minimizing
Eq.~\eqref{eq:rd_loss}, and quantize the decoder weights\;
\tcp{Stage 3: fine-tuning of the latents}
\For{\textnormal{each iteration}}{
  Round the latents with a straight-through gradient, and update only the
  latents with the quantized decoder weights fixed\;
}
Encode $\{\hat Y_i\}_{i=0}^{N}$ with the probabilities predicted by
$f_\varphi$, and write them to the bitstream together with the quantized
decoder weights $\varphi,\gamma,\theta$\;
\end{algorithm}

\subsection{Rate--Distortion Optimization}
\label{sec:objective}
All parameters of the codec, \ie{}, the latents and the three decoder
modules, are fitted jointly to each image. They are optimized for the
trade-off between the quality of the decoded sphere and the size of the
bitstream:
\begin{equation}
\mathcal{L} = D(\mathbf{x},\hat{\mathbf{x}})
  + \frac{\lambda}{HW}\bigl(\mathcal{R}_{\mathrm{latent}}
  + \mathcal{R}(\varphi,\gamma,\theta)\bigr),
\label{eq:rd_loss}
\end{equation}
where $D$ is the distortion measured on the sphere,
$\mathcal{R}_{\mathrm{latent}}$ is the latent rate of
Eq.~\eqref{eq:latent_rate}, $\mathcal{R}(\varphi,\gamma,\theta)$ is the
rate of the quantized decoder weights, and $\lambda$ controls the
trade-off between rate and distortion. Both rates are in bits, and the
division by $HW$ converts them to bits per pixel (bpp).

\textbf{Distortion.} Since the decoded image is defined on the sphere,
the distortion is also measured on the sphere, at a dense set of
$M_{\mathrm{s}}$ points $\{\mathbf{s}_m\}_{m=1}^{M_{\mathrm{s}}}$:
\begin{equation}
  D(\mathbf{x},\hat{\mathbf{x}})
  = \frac{1}{3M_{\mathrm{s}}}\sum_{m=1}^{M_{\mathrm{s}}}
    \bigl\|\mathbf{x}(\mathbf{s}_m)-\hat{\mathbf{x}}(\mathbf{s}_m)\bigr\|_2^2,
  \label{eq:distortion}
\end{equation}
where the points $\{\mathbf{s}_m\}$ form a Fibonacci lattice that covers
the sphere almost uniformly and differs from the viewport rays used for
evaluation, and $\mathbf{x}(\mathbf{s}_m)$ and
$\hat{\mathbf{x}}(\mathbf{s}_m)$ are interpolated from the ERP pixels and
from the decoded values at the nearest sampling points, respectively. Since the lattice is nearly uniform, the plain average in Eq.~\eqref{eq:distortion} approximates the squared error integrated over the sphere, without any latitude weighting.

\textbf{Decoder weight rate.} The decoder weights are part of the
bitstream, and their rate is the total length of their Exp-Golomb codes:
\begin{equation}
  \mathcal{R}(\varphi,\gamma,\theta)
  = \sum_{\alpha\in\{\varphi,\gamma,\theta\}}\sum_{\omega\in\alpha}
    \ell_\alpha\bigl(\lfloor\omega/\Delta_\alpha\rceil\bigr),
  \label{eq:param_rate}
\end{equation}
where $\ell_\alpha(\cdot)$ is the length of the Exp-Golomb codeword of
module $\alpha$, whose order is selected together with $\Delta_\alpha$.

\textbf{Training procedure.} We train the latents and the decoder
weights with the second-order optimizer SOAP~\cite{vyas2024soap} on
$D+\lambda\,\mathcal{R}_{\mathrm{latent}}/(HW)$. The decoder weights are
then quantized, with the step $\Delta_\alpha$ and the Exp-Golomb order of
each module selected by minimizing Eq.~\eqref{eq:rd_loss}. Finally, the
latents are fine-tuned with the quantized weights fixed. The details are
given in Algorithm~\ref{alg:training}.

%===========================================================================
\begin{figure*}[!t]
\centering
\includegraphics[width=\textwidth]{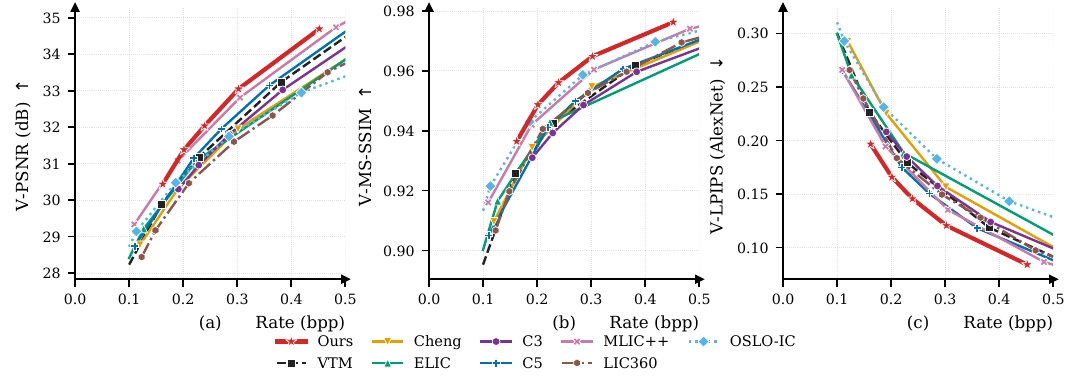}
\caption{Rate--distortion curves on the $100$-image benchmark under
V-PSNR (left), V-MS-SSIM (center), and V-LPIPS (right, lower is better),
computed on the six viewports of Sec.~\ref{sec:setup}. C5 denotes
Cool-Chic~5.0.}
\label{fig:rd_curves}
\end{figure*}

\begin{figure*}[!t]
\centering
\includegraphics[width=\textwidth]{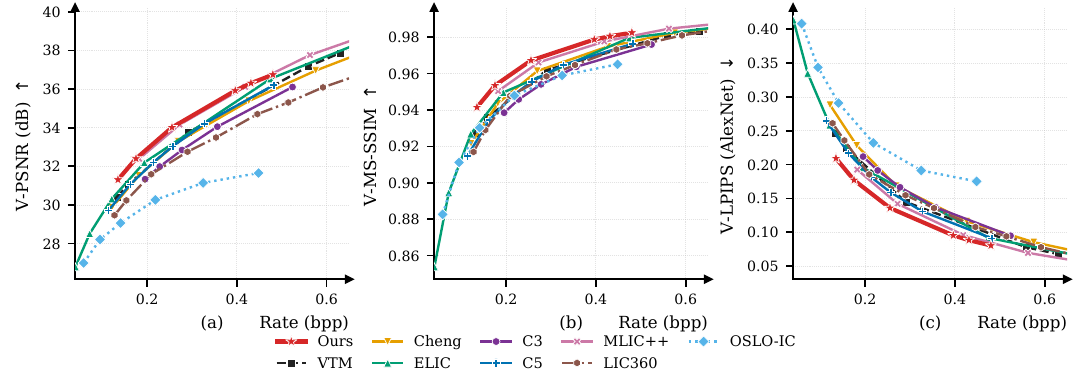}
\caption{Rate--distortion curves at $1536\times3072$ on the $99$ images
, computed on six $1024\times512$
viewports. LPIPS uses AlexNet, and C5 denotes Cool-Chic~5.0.}
\label{fig:highres_metrics}
\end{figure*}

\section{Experiments}
\label{sec:experiments}
In this section, we first describe the experimental setup. We then
compare our codec with traditional, learned, and INR codecs, both
quantitatively and visually, and finally report ablation studies and a
complexity analysis.

\subsection{Experimental Setup}
\label{sec:setup}

\textbf{Datasets.}
The main experiments use the 100-image omnidirectional benchmark of
Li~\etal~\cite{Li2024ContentPseudo} at a resolution of $768\times1536$. To
test a higher resolution, we further use 99 panoramas of size
$1536\times3072$ from the larger set released by
Li~\etal~\cite{Li2024ContentPseudo}. All results on both datasets are
evaluated with the viewport metrics described below.

\textbf{Evaluation protocol.}
We measure quality on six rectilinear viewports rendered from the
decoded sphere: four at the equator (yaw
$0^{\circ}/90^{\circ}/180^{\circ}/270^{\circ}$ at pitch $0^{\circ}$)
and two at the poles (pitch $\pm90^{\circ}$). Each viewport spans
$150^{\circ}$ horizontally and $60^{\circ}$ vertically, and has
$512\times256$ pixels in the main setting and $1024\times512$ pixels in
the high-resolution setting. The viewports are rendered in the same way for all codecs:
the reference and every reconstruction are read out along the same
viewing rays, and each reconstruction is read directly from its own
decoded format, \ie{}, a HEALPix sphere for our codec and OSLO-IC, and
an ERP image for the planar baselines. Each codec therefore goes
through exactly one mapping from its decoded format to the viewports.

\begin{table}[!t]
\centering
\caption{BD-rate(\%) of our codec against each
baseline under the three viewport metrics; negative values mean fewer
bits at the same quality. The left half uses the common range of
$0.1$--$0.5$\,bpp (Fig.~\ref{fig:rd_curves}), and the right half uses
the range shared by each pair of codecs (Fig.~\ref{fig:highres_metrics}).}
\label{tab:bdrate}
\footnotesize
\setlength{\tabcolsep}{3pt}
\resizebox{\columnwidth}{!}{%
\begin{tabular}{lcccccc}
\toprule
& \multicolumn{3}{c}{$768\times1536$ ($100$ images)}
& \multicolumn{3}{c}{$1536\times3072$ ($99$ images)} \\
\cmidrule(lr){2-4}\cmidrule(lr){5-7}
Ours vs. & V-PSNR & V-MS-SSIM & V-LPIPS & V-PSNR & V-MS-SSIM & V-LPIPS \\
\midrule
VVC~\cite{VVC}                            & $-16.3$ & $-26.5$ & $-19.5$ & $-15.8$ & $-24.2$ & $-19.0$ \\
Cheng~\cite{Cheng2020}                    & $-24.4$ & $-25.5$ & $-25.8$ & $-19.0$ & $-20.8$ & $-28.4$ \\
ELIC~\cite{elic}                          & $-19.3$ & $-25.3$ & $-23.3$ & $-10.5$ & $-16.3$ & $-18.8$ \\
C3~\cite{kim2024c3}                       & $-20.7$ & $-31.4$ & $-23.3$ & $-27.5$ & $-35.3$ & $-29.4$ \\
Cool-Chic~5.0~\cite{ladune2026coolchic5}  & $-13.1$ & $-24.9$ & $-13.9$ & $-18.7$ & $-27.6$ & $-18.8$ \\
MLIC++~\cite{jiang2023mlicpp}             &  $-5.5$ & $-12.5$ & $-12.8$ &  $-2.5$ &  $-9.4$ & $-10.6$ \\
LIC360~\cite{Li2022EndToEnd}              & $-28.6$ & $-26.0$ & $-20.3$ & $-32.8$ & $-29.3$ & $-24.0$ \\
OSLO-IC~\cite{Wawerek2025OSLOIC}          & $-27.8$ & $-12.2$ & $-42.0$ & $-44.7$ & $-41.2$ & $-43.0$ \\
\bottomrule
\end{tabular}}
\end{table}

\begin{table*}[!t]
\centering
\caption{Rate change (\%) of each ablation variant against our codec at
equal quality under the three viewport metrics; positive values mean more bits at the same quality. Each row of (a)--(c) changes a single component of our codec. MAC/px is the number of decoder multiplications per image pixel.}
\label{tab:ablation}
\footnotesize
\begin{tabular*}{\textwidth}{@{\extracolsep{\fill}}lcccc@{}}
\toprule
& & \multicolumn{3}{c}{Rate change at equal quality (\%)} \\
\cmidrule(lr){3-5}
Variant & MAC/px & V-PSNR & V-MS-SSIM & V-LPIPS \\
\midrule
Ours                                            & 4619 & ---     & ---     & ---     \\
\midrule
\multicolumn{5}{@{}l}{\textit{(a) Representation}} \\
HEALPix, 1 level                                & 2897 & $+44.5$ & $+52.5$ & $+28.6$ \\
ERP, 8 levels                                   & 6928 & $+4.2$ & $+6.6$ & $+5.6$ \\
ERP, 1 level                                    & 4346 & $+57.6$ & $+72.5$ & $+38.1$ \\
\midrule
\multicolumn{5}{@{}l}{\textit{(b) Entropy model}} \\
Inter-scale context: parent only                & 4619 & $+5.0$  & $+6.0$  & $+8.1$  \\
w/o inter-scale context                         & 3519 & $+2.3$  & $+2.9$  & $+3.8$  \\
Intra-scale context: $K$ previous in RING order & 4619 & $+7.1$  & $+5.4$  & $+4.7$  \\
No context                                      & 1585 & $+48.4$ & $+35.8$ & $+31.1$ \\
\midrule
\multicolumn{5}{@{}l}{\textit{(c) Spatial convolution}} \\
Multi-ring, one weight per ring                 & 4619 & $+1.5$  & $+1.5$  & $+5.3$  \\
Dilated kernel, $d=1,2,3$                       & 4022 & $+2.1$  & $+2.1$  & $+3.9$  \\
Multi-ring, $R_t=1,1,1$                         & 4022 & $+1.6$  & $+2.2$  & $+3.9$  \\
No spatial convolution                          & 3734 & $+7.2$  & $+4.5$  & $+4.3$  \\
\midrule
\multicolumn{5}{@{}l}{\textit{(d) Optimizer}} \\
SOAP $\rightarrow$ Adam                & 4619 & $+0.6$ & $+1.1$     & $+1.2$    \\
\bottomrule
\end{tabular*}
\end{table*}

\textbf{Metrics.}
We compare our codec with the baselines under V-PSNR,
V-MS-SSIM~\cite{wang2003msssim}, and V-LPIPS~\cite{zhang2018lpips}. For
every codec, the rate is the size of the bitstream actually produced
by that codec's own entropy coder. For the overfitted INR codecs,
including ours, which fit a model to each image, the bitstream contains
the coded latents, the quantized decoder weights, and the required
header data. For the learned codecs, whose networks are trained offline
and available at both the encoder and the decoder, the bitstream
contains only the coded latents.

\textbf{Compared codecs.}
We compare our codec with eight codecs: the traditional codec
VVC~\cite{VVC}, which serves as the anchor; the learned planar codecs
Cheng~\cite{Cheng2020}, ELIC~\cite{elic}, and
MLIC++~\cite{jiang2023mlicpp}; the planar INR codecs
C3~\cite{kim2024c3} and Cool-Chic~5.0~\cite{ladune2026coolchic5}; and
the learned omnidirectional codecs LIC360~\cite{Li2022EndToEnd} and
OSLO-IC~\cite{Wawerek2025OSLOIC}. The planar codecs and LIC360 are
applied directly to the ERP images. No trained model is released for
the OSLO-IC configuration, so we trained it ourselves from the authors'
released code. Moreover, OSLO-IC reports the distortion on its own
spherical sampling, so its published results cannot be compared
directly with those obtained under the JVET common test
conditions~\cite{jvet} (S-PSNR-NN~\cite{yu2015spsnr} and
WS-PSNR~\cite{sun2017wspsnr}) or with planar codecs evaluated on ERP.
To keep the comparison as fair as possible, we evaluate all
codecs in the viewport domain under the same protocol.

\subsection{Rate--Distortion Performance}
\label{sec:rd_results}

\textbf{Quantitative comparison.}
Fig.~\ref{fig:rd_curves} shows the rate--distortion curves under
V-PSNR, V-MS-SSIM, and V-LPIPS, all computed on the same six viewports, and
Table~\ref{tab:bdrate} lists the corresponding BD-rates. Compared with
the VVC anchor, our codec saves $16.3\%$, $26.5\%$, and $19.5\%$ of the
rate under V-PSNR, V-MS-SSIM, and V-LPIPS, respectively. Compared with
the learned planar codec, MLIC++, the savings are $5.5\%$,
$12.5\%$, and $12.8\%$. Compared with OSLO-IC, they are $27.8\%$,
$12.2\%$, and $42.0\%$.

\textbf{Higher resolution.}
Fig.~\ref{fig:highres_metrics} repeats the comparison at
$1536\times3072$, and the right half of Table~\ref{tab:bdrate} gives
the corresponding BD-rates. Our BD-rate remains negative against every
baseline under all three metrics, so the advantage of our codec holds
at this resolution.

\begin{figure*}[!t]
\centering
\includegraphics[width=\textwidth]{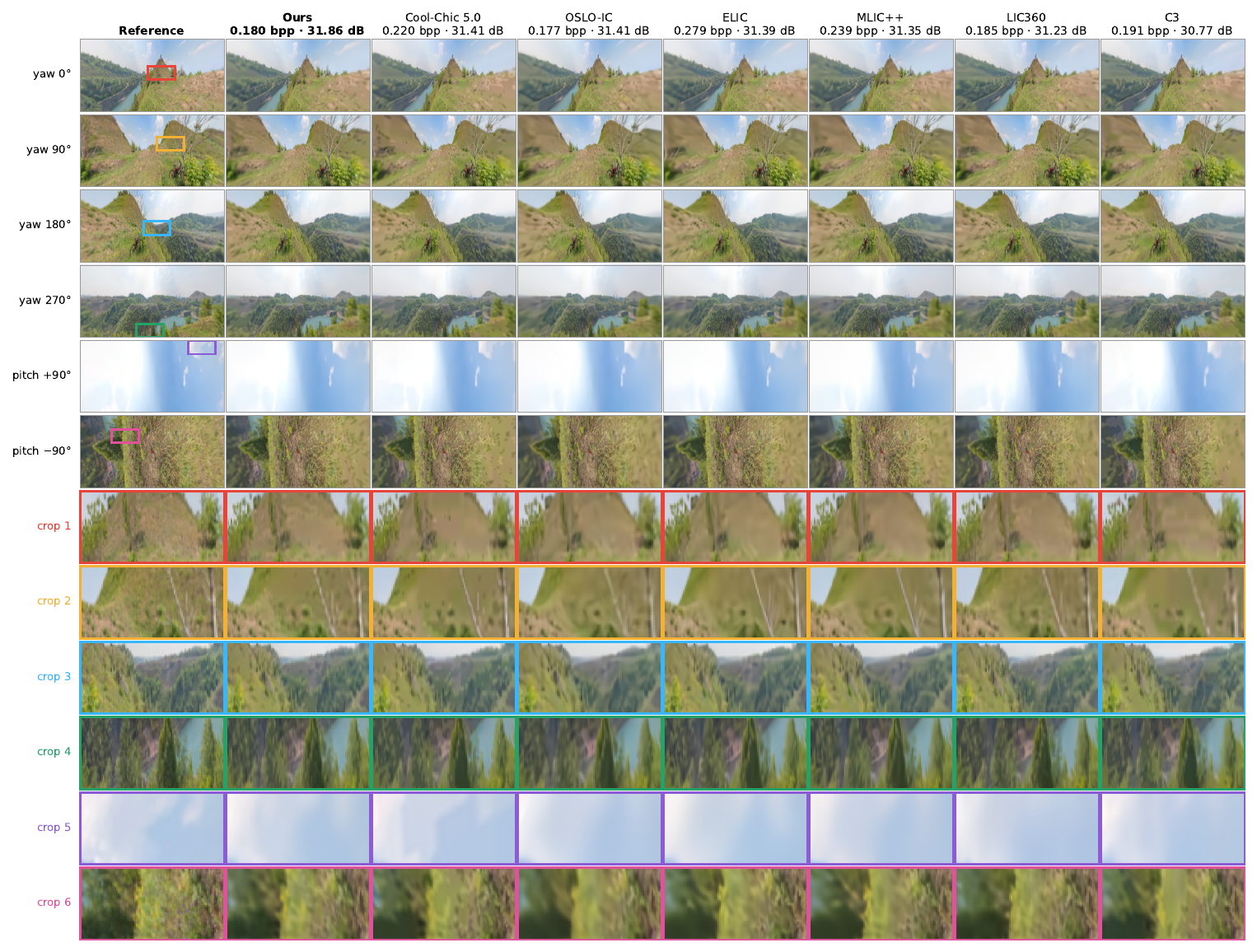}
\caption{\textbf{Visual comparison at similar rates on one panorama:
six viewports and six $5.3\times$ crops.} The columns show the
reference and seven codecs, sorted by V-PSNR. The crops are taken from
regions where our reconstruction is closer to the reference than all
baselines. Each baseline is shown at its operating point closest to
ours in rate (five of the six use $3$--$55\%$ more bits, and OSLO-IC
uses $1.7\%$ fewer), and all six have a V-PSNR $0.45$--$1.09$\,dB lower
than ours.}
\label{fig:qualitative}
\end{figure*}

\textbf{Visual comparison.}
Fig.~\ref{fig:qualitative} compares our codec with six baselines on
all six viewports of one panorama. Each baseline is shown at its
operating point closest to ours in rate: five of them use $3$--$55\%$
more bits than ours, and OSLO-IC uses $1.7\%$ fewer. All six have a
lower V-PSNR, by $0.45$--$1.09$\,dB.

\begin{figure*}[!t]
\centering
\includegraphics[width=\textwidth]{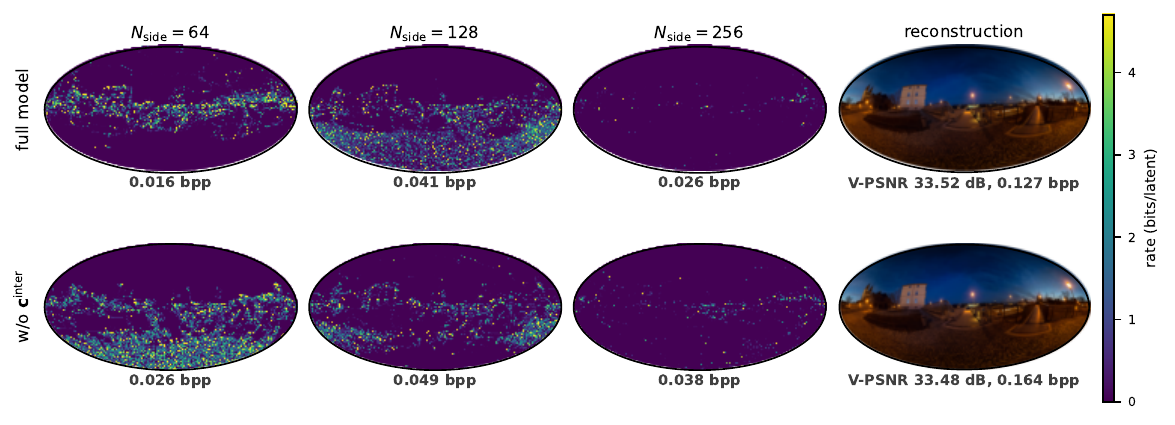}
\caption{Measured rate per latent for the three finest levels of one
panorama, for the full model (top) and for the same codec fitted without
the inter-scale context $\mathbf{C}^{\mathrm{Inter}}$ (bottom); the
number under each panel is the rate of that level in bpp, and the right
column shows the reconstruction with its V-PSNR and total rate. In both
rows, the rate follows the image structure rather than the latitude.
Without the inter-scale context, the finest level costs $46\%$ more
($0.026\rightarrow0.038$\,bpp), and the total rate rises by $29\%$
($0.127\rightarrow0.164$\,bpp) at a similar V-PSNR.}
\label{fig:rate_maps}
\end{figure*}

\subsection{Ablation Studies}
\label{sec:ablation}
We study four aspects of our codec in Table~\ref{tab:ablation}: the
latent representation, the entropy model, the spatial convolution, and
the optimizer. Each variant changes a single component of our codec. We
report its rate change at equal quality under the three viewport
metrics, together with the number of decoder multiplications per pixel (MAC/px).

\begin{figure}[!t]
\centering
\includegraphics[width=\columnwidth]{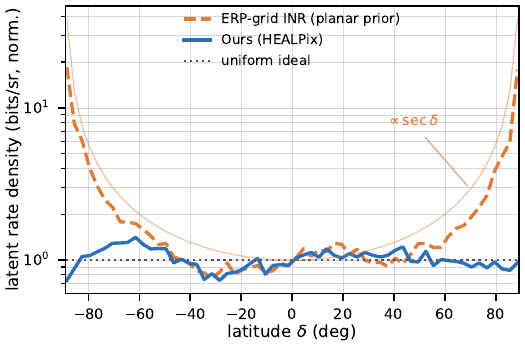}
\caption{\textbf{Areal bit density versus latitude}, \ie{}, bits per
unit area on the sphere normalized by the equatorial value, for
Cool-Chic~5.0 on ERP and for our codec on the test set. Toward the
poles, the ERP codec rises steeply, though below the $\sec\delta$ law of
Eq.~\eqref{eq:erp_density} (thin line), whereas ours stays near the
uniform ideal (dotted).}
\label{fig:rate_lat}
\end{figure}

\textbf{(a) Representation.} We replace the hierarchical HEALPix grids
by a single HEALPix level, or move the codec to ERP with eight levels or
a single level, keeping all other settings unchanged. A single level
costs $44.5\%$ more bits on HEALPix and $57.6\%$ more on ERP, making the
hierarchy the most important component: the coarse levels encode the
low-frequency structure at a much lower rate. With eight levels, ERP
still costs $4.2\%$ more bits than HEALPix and needs $6928$ instead of
$4619$ MAC/px, since its finest grid has $1.5$ times as many samples.
Equal-area sampling thus lowers both the rate and the decoding cost.

The grid also determines where the bits go. Fig.~\ref{fig:rate_lat}
compares the number of bits spent per unit area on the sphere, which we
call the areal bit density, for an ERP-grid overfitted codec
(Cool-Chic~5.0) and for our codec. To isolate the effect of the grid
from that of the entropy model, the rate of each latent is estimated
identically for both codecs, \ie{}, from the empirical frequency of its
quantized value within its level, without any context. Above
$|\delta|=60^{\circ}$, a region that covers $13.4\%$ of the sphere, our
codec spends $14.7\%$ of its bits, much closer to the area share than
the $27.9\%$ spent by the ERP codec, whose areal bit density reaches
about $18$ times the equatorial value near the poles. This confirms the
redundant polar rate of the ERP grid predicted by
Eq.~\eqref{eq:erp_density}.

\textbf{(b) Entropy model.} Without any context, the rate increases by
$48.4\%$. Taking the $K$ previous latents in RING order instead of the
$K$ nearest ones on the sphere costs $7.1\%$ more bits at the same
complexity, which confirms the benefit of selecting the neighbors by
their spherical distance. The inter-scale context reduces the rate by
$2.3\%$ at a cost of $1100$ MAC/px. As shown in
Fig.~\ref{fig:rate_maps}, it lowers the rate of the finest level, which
holds $75\%$ of all latents: with it, this level costs $0.026$\,bpp,
less than the $0.041$\,bpp of the next coarser level; without it, $46\%$
more. Keeping only the parent is even worse ($+5.0\%$), so the full
chain of ancestors is needed.

\textbf{(c) Spatial convolution.} Removing the spatial convolution of the
decoder increases the rate by $7.2\%$. At the same complexity ($4022$
MAC/px), a dilated kernel with $d=1,2,3$ and a single ring in all blocks
($R_t=1,1,1$) cost $2.1\%$ and $1.6\%$ more bits, respectively.
Enlarging the receptive field by dilation is thus less effective than
our multi-ring convolution, which covers every sampling point within
$R$ hops rather than distant samples that are weakly correlated with
the center. Sharing one weight per ring keeps the receptive field and
the complexity unchanged but costs $1.5\%$ more bits under V-PSNR and
$5.3\%$ more under V-LPIPS, which shows that the direction of each
neighbor matters for oriented structures such as edges.

\textbf{(d) Optimizer.} SOAP mainly accelerates convergence and shortens
the fitting time: under the same number of iterations, replacing it with
Adam increases the rate by $0.6\%$. Since the optimizer only affects the
fitting, the decoding complexity is unchanged.

\subsection{Complexity Analysis}
\label{sec:complexity}

Table~\ref{tab:complexity} lists the encoding and decoding times per
image. Decoding is timed on a single CPU core: every
codec runs on one core of an Intel Xeon Silver~4310 with batch size
$1$ and decodes a real bitstream with its own official entropy coder,
and model building and checkpoint loading are excluded from the
timing. The only exception is LIC360, whose released code runs only on
a GPU; its times, marked with~$\dagger$, are measured on one NVIDIA
RTX~4090 and are therefore not directly comparable with the others.

\begin{table}[!t]
\centering
\caption{Encoding and decoding time per image on the test set
($768\times1536$). Decoding times are measured on a single CPU core
with batch size~$1$. The encoding time of our codec is the per-image
fitting time (Algorithm~\ref{alg:training}) on one GPU. VVC is CPU-only
reference software. $^\dagger$LIC360 runs only on a GPU.}
\label{tab:complexity}
\setlength{\tabcolsep}{6pt}
\begin{tabular}{l c c c}
\toprule
Method & Paradigm & Enc.\ (s) $\downarrow$ & Dec.\ (s) $\downarrow$ \\
\midrule
VVC~\cite{VVC}         & Traditional & $160.080$ & $0.324$ \\
\midrule
Cheng~\cite{Cheng2020} & Autoencoder & $14.253$  & $27.910$ \\
ELIC~\cite{elic}       & Autoencoder & $15.662$  & $25.331$ \\
MLIC++~\cite{jiang2023mlicpp}       & Autoencoder & $16.893$  & $21.131$ \\
LIC360~\cite{Li2022EndToEnd}        & Autoencoder & $0.466^{\dagger}$  & $0.601^{\dagger}$ \\
OSLO-IC~\cite{Wawerek2025OSLOIC}    & Autoencoder & $22.004$  & $20.105$ \\
\midrule
C3~\cite{kim2024c3}    & Overfitted & $1000$     & $5.721$ \\
Cool-Chic 5.0~\cite{ladune2026coolchic5}     & Overfitted & $800$     & $5.994$ \\
\textbf{Ours}          & Overfitted & $1360$ & $7.674$ \\
\bottomrule
\end{tabular}
\end{table}
% ===========================================================================
\section{Conclusion and Discussion}
\label{sec:conclusion}
In this paper, we have presented an overfitted codec for
omnidirectional images that represents each image on the HEALPix sphere
instead of the ERP plane. Since all HEALPix pixels have equal area, the
latent grids distribute bits evenly over the sphere and avoid the polar
oversampling of ERP. Based on this representation, we designed a
spherical inter-scale conditional entropy model, which uses both
same-level neighbors and coarser-level ancestors as context, and a
multi-ring spherical convolution, which enlarges the receptive field of
the synthesis network while keeping the decoder small. Experiments with
viewport-based metrics show that our codec saves $16.3\%$, $26.5\%$, and
$19.5\%$ of the bit rate relative to VVC under V-PSNR, V-MS-SSIM, and
V-LPIPS, respectively, outperforms all compared codecs, and decodes
$2.6$--$3.6\times$ faster than the compared learned autoencoders.

Our codec still has some limitations. The entropy decoding within each
level is serial and dominates the decoding time, which could be reduced
by decoding groups of sampling points in parallel. In addition, the
input images are stored in ERP and must be resampled onto the sphere
before fitting. This resampling degrades the target and limits the
achievable quality, and its loss is included in all reported results.
Building omnidirectional datasets stored natively in HEALPix is
therefore part of our future work.

% ===========================================================================
% \section*{Acknowledgment}
% The authors would like to thank XXXX.

% ===========================================================================
\bibliographystyle{IEEEtran}
\bibliography{mainB}

\end{document}